\documentclass{bmvc2k}

\usepackage{amsmath,amsfonts,bm}

\def\eqref#1{equation~\ref{#1}}
\def\1{\bm{1}}

\def\vtheta{{\bm{\theta}}}

\def\ve{{\bm{e}}}

\def\vm{{\bm{m}}}

\def\vs{{\bm{s}}}

\def\vw{{\bm{w}}}
\def\vx{{\bm{x}}}

\def\vz{{\bm{z}}}

\DeclareMathAlphabet{\mathsfit}{\encodingdefault}{\sfdefault}{m}{sl}
\SetMathAlphabet{\mathsfit}{bold}{\encodingdefault}{\sfdefault}{bx}{n}

\def\sD{{\mathbb{D}}}
\DeclareMathOperator*{\argmin}{arg\,min}

\usepackage{hyperref}
\usepackage{url}
\usepackage{graphicx}
\usepackage[algo2e]{algorithm2e}
\usepackage{multirow}
\usepackage{xcolor}
\usepackage{mathtools}
\usepackage{relsize}
\usepackage{nicefrac}
\usepackage{amsmath,amssymb,enumerate}
\usepackage{amsthm,cancel}
\usepackage{enumitem}
\usepackage{dsfont}
\usepackage[utf8]{inputenc} % allow utf-8 input
\usepackage[T1]{fontenc}    % use 8-bit T1 fonts
\usepackage{microtype}      % microtypography
\usepackage{algorithm}
\usepackage{algorithmic}
\usepackage{animate}
\usepackage{multirow}

\usepackage{tikz}
\usetikzlibrary{shapes}

\usepackage{pifont}
\newcommand{\cmark}{\ding{53}}%
\newcommand{\xmark}{\ding{51}}%
\usepackage{booktabs}

\SetCommentSty{mycommfont}
\newcommand\blfootnote[1]{%
  \begingroup
  \renewcommand\thefootnote{}\footnote{#1}%
  \addtocounter{footnote}{-1}%
  \endgroup
}

\title{Overcoming Mode Collapse with Adaptive Multi Adversarial Training}

\addauthor{Karttikeya Mangalam*}{mangalam@cs.berkeley.edu}{1}
\addauthor{$\text{Rohin Garg*}^\dagger$}{rgarg@ucsd.edu}{2}
\addinstitution{
 University of California at Berkeley
}
\addinstitution{
 University of California at San Diego
}
\runninghead{Mangalam, Garg}{Adaptive Multi Adversarial Training}

\begin{document}

\maketitle
\vspace{-1mm}
\begin{abstract}
Generative Adversarial Networks (GANs) are a class of generative models used for various applications, but they have been known to suffer from the \textit{mode collapse} problem, in which some modes of the target distribution are ignored by the generator. Investigative study using a new data generation procedure indicates that the mode collapse of the generator is driven by the discriminator's inability to maintain classification accuracy on previously seen samples, a phenomenon called Catastrophic Forgetting in continual learning. Motivated by this observation, we introduce a novel training procedure that adaptively spawns additional discriminators to remember previous modes of generation. On several datasets, we show that our training scheme can be plugged-in to existing GAN frameworks to mitigate mode collapse and improve standard metrics for GAN evaluation. Code and pre-trained models are available at \href{https://github.com/gargrohin/AMAT}{https://github.com/gargrohin/AMAT}\vspace{-4mm}
\end{abstract}

%-------------------------------------------------------------------------
\section{Introduction}
\vspace{-2mm}
\label{sec:intro}
Generative Adversarial Networks (GANs) \citep{goodfellow2014generative} are an extremely popular class of generative models used for text and image generation in various fields of science and engineering, including biomedical imaging~\citep{yi2019generative,nie2018medical,wolterink2017generative}, autonomous driving~\citep{hoffman2018cycada,zhang2018deeproad}, and robotics~\citep{rao2020rl,bousmalis2018using}. However, GANs are widely known to be prone to \textit{mode collapse}, which refers to a situation where the generator only samples a few modes of the real data, failing to faithfully capture other more complex or less frequent categories. While the mode collapse problem is often overlooked in text and image generation tasks, and even traded off for higher realism of individual samples~\citep{karras2019style,brock2018large}, dropping infrequent classes may cause serious problems in real-world problems, in which the infrequent classes represent important anomalies. For example, a collapsed GAN can produce racial/gender biased images~\citep{Menon_2020_CVPR}.\blfootnote{* Equal technical contribution. $\dagger$ Work done during internship.}

\begin{figure*}
\centering
\includegraphics[trim={0 0 0 10},clip,width=0.90\textwidth]{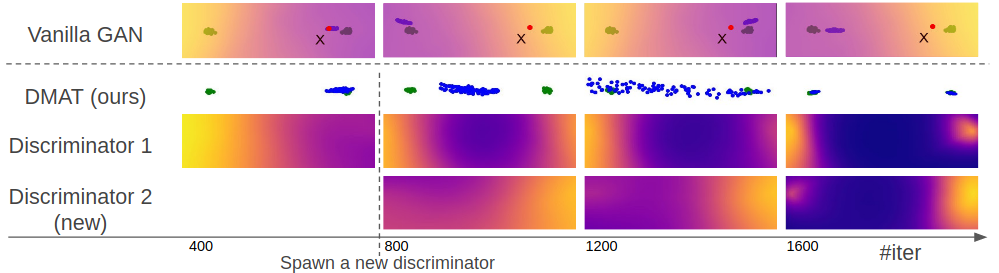}
\vspace{3mm}
\caption{\textbf{Visualizing training trajectories}: Distribution of real (green dots) and fake (blue dots) over the course of vanilla GAN (top row) and our method (the second row and below). The background color indicates the prediction heatmap of the discriminator with blue being fake and warm yellow being real. Once the vanilla GAN falls into mode collapse (top row), it ends up oscillating between the two modes without convergence. Also, the discriminator's prediction at point X oscillates, indicating catastrophic forgetting in the discriminator. AMAT algorithm adapts to the need, and a new discriminator is spawned during training which effectively learns the forgotten mode, guiding the GAN optimization toward convergence. } 
\label{fig:oscillation}
\vspace{-3mm}
\end{figure*}

Moreover, mode collapse causes instability in optimization, which can damage both diversity and the realism of individual samples of the final results. As an example, we visualized the training progression of the vanilla GAN~\citep{goodfellow2014generative} for a simple bimodal distribution in the top row of Figure~\ref{fig:oscillation}. At collapse, the discriminator conveniently assigns high realism to the region unoccupied by the generator, regardless of the true density of real data. This produces a strong gradient for the generator to move its samples toward the dropped mode, swaying mode collapse to the other side. So, the discriminator loses its ability to detect fake samples it was previously able to, such as point \textbf{X}\tikz\draw[red,fill=red] (0,5) circle (.3ex);. The oscillation continues without convergence.

We observe that the mode collapse problem is closely related to Catastrophic Forgetting~\citep{mccloskey1989catastrophic, mcclelland1995there, ratcliff1990connectionist} in continual learning. 
%That is, since the distribution of the generated samples is not stationary, the discriminator \textit{forgets} to classify the previously generated samples as fake, hindering convergence of the GAN minimax game.
A promising line of works~\citep{sidetuning2019, NIPS2019_9429, rusu2016progressive, fernando2017pathnet} tackle the problem in the supervised learning setting by instantiating multiple predictors, each of which takes charge in a particular subset of the whole distribution. We also tackle the problem of mode collapse in GAN by tracking the severity of Catastrophic Forgetting by storing a few exemplar data during training, spawning an additional discriminator if forgetting is detected, Figure~\ref{fig:oscillation}. The key idea is that the added discriminator is left intact unless the generator recovers from mode dropping of that sample, essentially sidestepping catastrophic forgetting. 

%This \textit{sidesteps} the problem of catastrophic forgetting since a different set of weights are responsible for different parts of the data distribution. Likewise, we also tackle the mode collapsing problem by implicitly dividing the entire generated distribution into disjoint partitions, and training a separate discriminator for each partition. Because the distribution is more stationary at each partition, the discriminators are more robust to catastrophic forgetting. %Note that such partitioning is a nontrivial task, as the accumulated distribution of the generated samples over training is not known a priori. 
We show that our proposed approach based on adaptive addition of discriminators can be added to any of the existing GAN frameworks, and is most effective in preventing mode collapse. Furthermore, the improved stability of training boosts the standard metrics on popular GAN frameworks. To summarize, our contributions are: \emph{First}, we propose a novel GAN framework, named Adaptive Multi Adversarial Training (AMAT), that effectively prevents Catastrophic Forgetting in GANs by spawning additional discriminators during training. \emph{Second}, we also propose a computationally efficient synthetic data generation procedure for studying mode collapse in GANs that allows visualizing high dimensional data using normalizing flows. We show that mode collapse occurs even in the recent robust GAN formulations. \emph{Third}, our method can be plugged into any state-of-the-art GAN frameworks and still improve the quality and coverage of the generated samples.

% \begin{itemize}
%     \item We propose a novel GAN framework, named Adaptive Multi                                       Adversarial Training (AMAT), that effectively prevents Catastrophic Forgetting in GANs by spawning additional discriminators during training.
%     \item We propose a computationally efficient synthetic data generation procedure for studying mode collapse in GANs that allows visualizing high dimensional data using normalizing flows. We show that mode collapse occurs even in the recent robust GAN formulations.
%     \item Our method can be plugged into any state-of-the-art GAN frameworks and still improve the quality and coverage of the generated samples.
% \end{itemize}

\vspace{-5mm}
\section{Related Works}
\vspace{-2mm}
\label{sec:related}
Previous works have focused on independently solving either catastrophic forgetting in supervised learning or mode collapse during GAN training. In this section we review these works in detail and discuss our commonalities and differences.  
\vspace{-2mm}
\subsection{Mitigating Mode Collapse in GANs}
Along with advancement in the perceptual quality of images generated by GAN~\citep{miyato2018spectral,karras2019style,brock2018large,karras2020analyzing}, a large number of papers~\citep{durugkar2016generative,metz2016unrolled,arjovsky2017wasserstein,srivastava2017veegan,nguyen2017dual,lin2018pacgan,MeschederICML2018,karras2019style} identify the problem of mode collapse in GANs and aim to mitigate it. However mode collapse was seen as a secondary symptom that would be naturally solved as the stability of GAN optimization progresses~\citep{arjovsky2017wasserstein,MeschederICML2018,bau2019seeing}. 
%While the magnitude of mode collapse is certainly mitigated with more stable optimization, we show that it is still not a solved problem. 
To explicitly address mode collapse, Unrolled GAN~\citep{metz2016unrolled} proposes an unrolled optimization of the discriminator to optimally match the generator objective, thus preventing mode collapse. VEEGAN~\citep{srivastava2017veegan} utilizes the reconstruction loss on the latent space. PacGAN~\citep{lin2018pacgan} feeds multiple samples of the same class to the discriminator when making the decisions about real/fake. In contrast, our approach can be plugged into existing state-of-the-art GAN frameworks to yield additional performance boost.
\vspace{-4mm}
\subsection{Multi-adversarial Approaches}
\vspace{-1mm}
The idea of employing more than one adversarial network in GANs to improve results has been explored by several previous works independent of the connection to continual learning and catastrophic forgetting. MGAN~\citep{hoang2018mgan} uses multiple generators, while D2GAN~\citep{nguyen2017dual} uses two discriminators, and GMAN~\citep{durugkar2016generative} and MicrobatchGAN~\citep{mordido2020microbatchgan} proposed a method with more than two discriminators that can be specified as a training hyperparameter beforehand. 
%Using multiple discriminators is one such additive method to tackle catastrophic forgetting. D2GAN \citep{nguyen2017dual} propose two discriminator training, and \citep{durugkar2016generative} introduce GANs with multiple discriminators. Both methods target mode collapse specifically, but neither of them address catastrophic forgetting. Microbatch GAN~\citep{durugkar2016generative} forces the discriminators to learn separate modes implicitly through, however it does not lead to significant performance gains. 
However, all previous works require the number of discriminators to be fixed beforehand, which is a major drawback since it depends on several intricate factors such as training dynamics, data distribution complexity, model architecture, initialization hyper-parameters etc. and is expensive and difficult to approximate even with several runs of the algorithm. In contrast, noting by the connection of multi-adversarial training to parameter expansion approaches to catastrophic forgetting, we propose an \textit{adaptive} method that can add discriminators incrementally during training thus achieving superior performance than existing works both on data quality metrics as well as overall computational effort. 

\begin{table*}[t]
\resizebox{\textwidth}{!}{
\begin{tabular}{c|c|c|c|c|c||c}
\begin{tabular}{c}
$g(\mathbf{z}) = $ 
\end{tabular} 
& ${1}$ 
& \begin{tabular}[c]{@{}c@{}} 
$\mathbf{A}_{392 \times 2}$ 
\\ \end{tabular} & 
\begin{tabular}[c]{@{}c@{}}
$\mathbf{z}$
\end{tabular} 
& \begin{tabular}[c]{@{}c@{}}
MLP 
\end{tabular} 
& \begin{tabular}[c]{@{}c@{}}
MLP, $\mathbf{A}_{392 \times 2}$ 
\end{tabular} 
& \begin{tabular}{@{}c@{}}
\small{MNIST} \end{tabular} 
\\[1ex]
\hline
\begin{tabular}[c]{@{}c@{}}
Label\\ \end{tabular} 
& {\fontfamily{lmtt}\selectfont Level I}      
& {\fontfamily{lmtt}\selectfont Level II}
& {\fontfamily{lmtt}\selectfont Level III}
& {\fontfamily{lmtt}\selectfont Level IV}
& {\fontfamily{lmtt}\selectfont Level V}
& -
\\ 
\hline
\begin{tabular}[c]{@{}c@{}}
\small{GAN-NS \citep{goodfellow2014generative}}\\ \end{tabular} 
& \cmark \enskip \big\vert \enskip  \xmark     
& \cmark \enskip \big\vert \enskip \xmark 
& \cmark \enskip \big\vert \enskip \cmark 
& \cmark \enskip \big\vert \enskip  \cmark 
& \cmark \quad   \big\vert \quad   \cmark
& \xmark 
\\ 
\begin{tabular}[c]{@{}c@{}}
\small{WGAN} \citep{arjovsky2017wasserstein} \\ \end{tabular}    
& \xmark \enskip \big\vert \enskip \xmark      
& \cmark \enskip \big\vert \enskip \xmark
& \cmark \enskip \big\vert \enskip \xmark
& \cmark \enskip \big\vert \enskip   \cmark  
& \cmark \quad   \big\vert \quad   \cmark 
& \xmark 
\\ 
\begin{tabular}[c]{@{}c@{}}
\small{Unrolled GAN} \citep{metz2016unrolled} \end{tabular}
&  \xmark \enskip \big\vert \enskip \xmark
&  \xmark \enskip \big\vert \enskip \xmark 
&  \xmark \enskip \big\vert \enskip \xmark 
&  \xmark \enskip \big\vert \enskip \xmark 
&  \cmark \quad   \big\vert \quad   \cmark    
&  \xmark 
\\
% \end{tabular}

\begin{tabular}[c]{@{}c@{}}
\small{D2GAN} \citep{nguyen2017dual} \end{tabular}
&  \xmark \enskip \big\vert \enskip \xmark
&  \xmark \enskip \big\vert \enskip \xmark
&  \xmark \enskip \big\vert \enskip \xmark
&  \xmark \enskip \big\vert \enskip \xmark
&  \cmark \quad   \big\vert \quad   \cmark
&  \xmark 
\\
% \end{tabular}

\begin{tabular}[c]{@{}c@{}}
\small{GAN-NS + AMAT}  \end{tabular}
&  \xmark \enskip \big\vert \enskip \xmark
&  \xmark \enskip \big\vert \enskip \xmark
&  \xmark \enskip \big\vert \enskip \xmark
&  \xmark \enskip \big\vert \enskip \xmark
&  \cmark \quad   \big\vert \quad   \cmark
&  \xmark 
\\
\hline
\end{tabular}
}
% \hline
\vspace{2mm}
\caption{\xmark \hspace{0.1mm} indicates that the generator could effectively learn all the data modes, while \cmark \hspace{0.1mm} means \textit{despite best efforts with tuning} the training suffers from mode collapse (more than a quarter of data modes dropped). We show results with the SGD (left) \& ADAM (right) optimizers. MNIST results with ADAM optimizer are provided for reference. We observe that MNIST is a relatively easy dataset, falling between {\fontfamily{lmtt}\selectfont Level I} and {\fontfamily{lmtt}\selectfont II} in terms of complexity.}
\label{tab:synthetic}
\vspace{-4mm}
\end{table*}

\vspace{-4mm}
\subsection{Overcoming Catastrophic Forgetting in GAN}
\vspace{-1mm}
%Catastrophic forgetting was first observed in connectionist networks by~\cite{mccloskey1989catastrophic}. Since then, a plethora of works has proposed solutions to mitigate catastrophic forgetting in neural networks. 
Methods to mitigate catastrophic forgetting can be categorized into three groups: a) regularization based methods~\citep{kirkpatrick2017overcoming} b) memory replay based methods~\citep{rebuffi2017icarl} c) network expansion based methods~\citep{zhang2018deeproad, NIPS2019_9429}. Our work is closely related to the third category of methods, which dynamically adds more capacity to the network, when faced with novel tasks. This type of methods, adds \emph{plasticity} to the network from new weights (fast-weights) while keeping the \emph{stability} of the network by freezing the past-weights (slow-weights). 
Additionally, we enforce stability by letting a discriminator to focus on a few set of classes, not by freezing its weights. 

The issue of catastrophic forgetting in GANs has been sparsely explored before. %with few works approaching the problem from disparate directions .
\citet{chen2018self} and \citet{tran2019self} propose a self-supervised learning objective to prevent catastrophic forgetting by adding new loss terms. \citet{liang2018generative} proposes an online EWC based solution to tackle catastrophic forgetting in the discriminator. We propose a prominently different approach based on parameter expansion rather than regularization. While the regularization based approaches such as \citet{liang2018generative} attempt to retain the previously learnt knowledge by constrained weight updates, the parameter expansion approaches effectively sidestep catastrophic forgetting by freezing previously encoded knowledge. \citet{thanhcatastrophic} also discuss the possibility of catastrophic forgetting in GAN training but their solution is limited to theoretical analyses with simplistic proposals such as assigning larger weights to real samples and optimizing the GAN objective with momentum. Practically, we observed that their method performs worse than a plain vanilla DCGAN on simple real world datasets like CIFAR10. In contrast, our method leverages insights from continual learning and has a direct connections to prevalent parameter expansion approaches in supervised learning. We benchmark extensively on several datasets and state-of-the-art GAN approaches where our method consistently achieves superior results to the existing methods.
%that tackle mode collapse in GANs.
\vspace{-7mm}
\section{Proposed Method}
\vspace{-2mm}
In this section, we first describe our proposed data generation procedure that we use as a petri dish for studying mode collapse in GANs. 
The procedure uses random normalizing flows for simultaneously allowing training on complex high dimensional distributions yet being perfectly amenable to 2D visualizations. 
Next, we describe our proposed Adaptive Multi Adversarial Training (AMAT) algorithm that effectively detects catastrophic forgetting and spawns a new discriminator to prevent mode collapse.
\vspace{-5mm}
\subsection{Synthetic Data Generation with Normalizing flows}
\vspace{-1mm}
Mode dropping in GANs in the context of catastrophic forgetting of the discriminator is a difficult problem to investigate using real datasets. This is because the number of classes in the dataset cannot be easily increased, the classes of fake samples are often ambiguous, and the predictions of the discriminator cannot be easily visualized across the whole input space. 
In this regard, we present a simple yet powerful data synthesis procedure that can generate complex high dimensional multi-modal distributions, yet maintaining perfect 2-D visualization capabilities. Samples from a 2-D Gaussian distribution are augmented with biases and subjected to an invertible normalizing flow~\citep{karami2019invertible} parameterized by well conditioned functions $g_i: \mathbb{R}^{d^0_i} \rightarrow \mathbb{R}^{d^1_i}$. This function can be followed by a linear upsampling transformation parameterized by a $d^1_i \times d^0_{i+1}$ dimensional matrix $A^i$ (Algorithm \ref{algo:synthetic}). 
  The entire transform is deliberately constructed to be a bijective function so that every generated sample in $\hat{y} \in \mathbb{R}^D$ can be analytically mapped to $\mathbb{R}^2$, allowing perfect visualization on 2D space. Furthermore, by evaluating a dense grid of points in $\mathbb{R}^2$, we can understand discriminator's learned probability distribution on $\mathbf{z}$ manifold as a heatmap on the 2D plane. 
  This synthetic data generation procedure enables studying mode collapse in a controlled setting. This also gives practitioners the capability to train models on a chosen data complexity with clean two-dimensional visualizations of both the generated data and the discriminator's learnt distribution. 
  This tool can be used for debugging new algorithms using insights from the visualizations. 
  In the case of mode collapse, a quick visual inspection would give the details of which modes face mode collapse or get dropped from discriminator's learnt distribution.
 \vspace{-2mm}
\subsection{Adaptive Multi Adversarial Training}
\vspace{-1mm}
Building upon the insight on relating catastrophic forgetting in discriminator to mode collapse in generator, we propose a multi adversarial generative adversarial network training procedure. 
The key intuition is that the interplay of catastrophic forgetting in the discriminator with the GAN minimax game, leads to an oscillation generator. Thus, as the generator shifts to a new set of modes the discriminator forgets the learnt features on the previous modes. 
\begin{minipage}[b]{.46\textwidth}
\begin{algorithm}[H]
\DontPrintSemicolon
% \SetAlgoLined
\SetNoFillComment
% \LinesNotNumbered 
% \SetAlgoLined
\begin{algorithmic}
\STATE \textbf{Input:} Mean $\{\mu_i\}_{i=1}^K$ and standard deviation $\{\sigma_i\}_{i=1}^K$  for initialization,  $\{g_i\}_{i=1}^{L}$ well conditioned $\mathbb{R}^2 \rightarrow \mathbb{R}^2$ functions  
\STATE Sample weights $\vw \sim \text{Dirichlet}(K)$ \\
\tcc{Sample from 2D gaussian mixture}
\STATE $\mathbf{x}_{2D} \sim \Sigma_{i = 1}^N w_i \mathcal{N}( \mu_i, \sigma_i)$  
\STATE $\mathbf{x}^0_{2D} = \Big[[x^0_{2D}; {1}], [x^1_{2D}; {1}] \Big] $ \newline
\tcc*{Randomly Init Normalizing Flow}
\FOR{$k = 1$ to $k = K$}
    % \newline \tcc*{Alternate Identity mapping} \newline
    \IF{k is even} 
    \STATE $\vx^k = \big[\vx^{k}_0, \vx^{k}_1 \cdot  g_k(\vx^k_0)\big] $  
    \ELSE
    % \IF{k is odd} 
    \STATE $\vx^k = \big[\vx^{k}_0 \cdot  g_k(\vx^k_1), \vx^{k}_1 \big] $  
    \ENDIF
\ENDFOR
\end{algorithmic}
\caption{Synthetic Data Generation}
\label{algo:synthetic}
\end{algorithm}
\vspace{-4mm}
% \end{minipage}
\begin{algorithm}[H]
\DontPrintSemicolon
% \SetAlgoLined
\SetNoFillComment
% \LinesNotNumbered 
% \SetAlgoLined
\caption{DSPAWN: Discriminator \mbox{Spawning} Routine}
\label{algo:spawn}
\begin{algorithmic}
\STATE \textbf{Require:} Exemplar Data $\{\ve\}_{i=1}^m$ 
\STATE \textbf{Input:} Discriminator set $\displaystyle \sD = \{f_w^{i}\}_{i=1}^K$ \newline \tcc*{Check forgetting on exemplars}
\FOR{$i = 1$ to $i = m$}
\STATE $\vs[k] \leftarrow f_w^{k}(\ve_i) \ \forall\  k\  \in \{1 \dots K\}$\;
    \IF{ $K *\max(\vs) > \alpha_t * \sum_k\vs[k]$}
         \STATE Initialize $f_w^{K+1}$ with random weights $w$ \newline \tcc*{Spawn a new discriminator} 
         \STATE Initialize random weight $w^{K+1}$
         \STATE $\displaystyle \sD \leftarrow \{f_w^{i}\}_{i=1}^K \bigcup f_w^{K+1}$
         \STATE \textbf{break}
    \ENDIF
\ENDFOR
\STATE \textbf{return} Discriminator Set $\displaystyle \sD$
\end{algorithmic}
\end{algorithm}
\end{minipage}\hfill
% \end{minipage}\hfill% This must go next to `\end{minipage}`
\begin{minipage}[b]{.48\textwidth}
\begin{algorithm}[H]
\caption{A-MAT: Adaptive Multi-Adversarial Training}
\label{algo:multi}
\DontPrintSemicolon
% \SetAlgoLined
% \SetNoFillComment
% \LinesNotNumbered 
\begin{algorithmic}
\STATE \textbf{Require: }$\vw^{i}_0$, $\vtheta_0$  initial discriminator \& generator params, greediness param $\epsilon$, $\{T_k\}$ spawn warmup iteration schedule

\STATE $\displaystyle \sD \leftarrow \{f_w^{0}\}$\;
\WHILE{$\vtheta$ has not converged}
  \STATE Sample $\{\vz^{(i)}\}_{i=1}^B \sim p(z)$
  \STATE Sample $\{\vx^{(i)}\}_{i=1}^B \sim \mathbb{P}_r$ 
  \STATE Sample $\{\sigma_1(i)\}_{i=1}^B \sim \text{Uniform}(1, K)$
  \STATE Sample $\{\alpha(i)\}_{i=1}^B \sim \text{Bernoulli}(\epsilon)$ \newline \tcc{Loss weights over discriminators}
  \STATE Sample weights $\vm \sim \text{Dirichlet}(K)$ 
  \STATE $\hat\vx^{(i)} \leftarrow g_\theta(\vz^{(i)}) $  
  \STATE $\sigma_2(i) \leftarrow \argmin_{k}f_w^{k}(\hat\vx^{(i)})$ \newline \tcc*{Discriminator responsible for $\hat{x}^{(i)}$}
  \STATE $\sigma_{z}(i) \leftarrow \alpha(i) \sigma_1(i) + (1 - \alpha(i)) \sigma_2(i)$ \newline \tcc*{Discriminator responsible for $x^{(i)}$} 
  \STATE $\sigma_x(i) \leftarrow \sigma_1(i)$ \newline \tcc*{Training Discriminators}
    $L_{w} \leftarrow \sum_{i=1}^B [f_w^{\sigma_{x}(i)}(\vx_i) - 1]^{-} -  [f_w^{\sigma_{z}(i)}(\hat\vx_i) + 1]^+$
  \FOR{$k = 1$ to $k = \vert \displaystyle \sD \vert$}
    \STATE $w^k \leftarrow \text{ADAM}(L_{w})$
  \ENDFOR \newline \tcc*{Training Generator}
  \STATE $s[k] \leftarrow \sum_{i=1}^B f_w^{k}(\hat\vx^{(i)}) \ \forall\  k\  \in \{1 \dots \vert \displaystyle \sD \vert\}$\newline
%   sorted\_indices $\leftarrow$ argsort(D\_score)\;
 \tcc*{Weighed mean over discriminators}
  \STATE $L_{\theta} \leftarrow \text{sort}(\vm) \cdot \text{sort}(s) $
  \STATE $\theta \leftarrow \text{ADAM}(L_{\theta})$\;
%   loss $\leftarrow  -\E_{\vz}[ \sum_{K}\text{weights}[k]*D_{\text{sorted\_indices}[k]}(G(\vz)) ]
    % \tcp*{Check condition to add $D_{K+1}$}
    \IF{more than $T_t$ warm-up iterations since the last spawn}
        \STATE $\displaystyle \sD \leftarrow \text{DSPAWN}(\{f_w^{i}\})$
    \ENDIF
\ENDWHILE
\end{algorithmic}
\end{algorithm}
\vspace{-3mm}
\end{minipage}
However if there are multiple discriminators available, each discriminator can implicitly \emph{specialize} on a subset of modes. Thus even if the generator oscillates, each discriminator can remember their own set of modes, and they will not need to move to different set of modes. This way we can effectively \emph{sidestep} forgetting and ensure the networks do not face significant distribution shift. A detailed version of our proposed method is presented in Algorithm \ref{algo:multi}.
\newline
\newline
\textbf{Spawning new discriminators}: We initialize the AMAT training Algorithm~\ref{algo:multi} with a regular GAN using just one discriminator. We also sample a few randomly chosen exemplar data points with a maximum of one real sample per mode, depending on dataset complexity. The exemplar data points are used to detect the presence of catastrophic forgetting in the currently active set of discriminators $\displaystyle \sD$ and spawn a new discriminator if needed. Specifically (Algorithm \ref{algo:spawn}), we propose that if \emph{any} discriminator among $\displaystyle \sD$ has an unusually high score over an exemplar data point $\ve_i$, this is because the mode corresponding to $\ve_i$ has either very poor generated samples or has been entirely dropped. 

In such a situation, if training were to continue we risk catastrophic forgetting in the active set $\displaystyle \sD$, if the generator oscillate to near $\ve_i$. This is implemented by comparing the $\max$ score over  $\displaystyle sD$ at $\ve_i$ to the average score over $\displaystyle \sD$ and spawning a new discriminator when the ratio exceeds $\alpha_t (> 1)$. Further, we propose to have $\alpha_t (> 1)$ a monotonically increasing function of $| \displaystyle \sD|$, thus successively making it harder to spawn each new discriminator. Additionally, we use a warm-up period $T_t$ after spawning each new discriminator from scratch to let the spawned discriminator train before starting the check over exemplar data-points.  
\newline
\newline
\noindent \textbf{Multi-Discriminator Training: }  We evaluate all discriminators in $\displaystyle \sD$ on the fake samples but do not update all of them for all the samples. Instead, we use the discriminator scores to assign responsibility of each data point to only one discriminator. 
\newline
\newline
\noindent \textbf{Training over fake samples}: We use an $\epsilon$-greedy approach for fake samples where the discriminator with the lowest output score is assigned responsibility with a probability $1 - \epsilon$ and a random discriminator is chosen with probability $\epsilon$. 
\newline
\newline
\noindent \textbf{Training over real samples}: The discriminator is always chosen uniformly randomly thus we slightly prefer to assign the same discriminator to the fake datapoints from around the same mode to ensure that they do not forget the already learnt modes and switch to another mode. The random assignment of real points ensure that the same preferentially treated discriminator also gets updated on real samples. 

Further for optimization stability, we ensure that the real and fake sample loss incurred by each discriminator is roughly equal in each back-propagation step by dynamically reweighing them by the number of data points the discriminator is responsible for. We only update the discriminator on the losses of the samples they are responsible for. 
While it may seem that adding multiple discriminators makes the procedure expensive, in practice the total number of added discriminator networks never surpass three for the best results.
%addition
It is possible to change the hyperparameters to allow a large number of discriminators but that results in sub-optimal results and incomplete training. The optimal hyperparameter selection is explained in the Appendix for each dataset. %\ref{appendix:cifarexp}.
Further, the additional discriminators get added during later training stages and are not trained from the start, saving compute in comparison to prior multi-adversarial works which train all the networks from the beginning. 
Also, unlike AdaGAN \citep{tolstikhin2017adagan} and similar Boosted GAN models that need to store multiple Generators post training,  the final number of parameters required during inference remains unchanged under AMAT . Thus the inference time remains the same, but with enhanced mode coverage and sample diversity. Unlike \citep{tolstikhin2017adagan}, our discriminator addition is adaptive, i.e. discriminators are added during the training thus being more efficient.
\newline
\newline
\noindent \textbf{Generator Training:} We take a weighted mean over the discriminators scores on the fake datapoints for calculating the generator loss. At each step, the weights each discriminator in $\displaystyle \sD$ gets is in decreasing order of its score on the fake sample. Hence, the discriminator with the lowest score is given the most weight since it is the one that is currently specializing on the mode the fake sample is related to. In practice, we sample weights randomly from a Dirichlet distribution (and hence implicitly they sum to $1$) and sort according to discriminator scores to achieve this. We choose soft weighing over hard binary weights because since the discriminators are updated in an $\epsilon$ greedy fashion, the discriminators other than the one with the best judgment on the fake sample might also hold useful information. Further, we choose the weights randomly instead of fixing a chosen set to ensure AMAT is more deadset agnostic since the number of discriminator used changes with the dataset complexity so does the number of weights needed. While a suitable function for generating weights can work well on a particular dataset, we found random weights to work as well across different settings.  
\vspace{-3mm}
\section{Results}
\vspace{-3mm}
\label{sec:results}

\begin{figure*}[t!]
    \centering
    \begin{subfigure}%{0.48\textwidth}
        \centering
        \includegraphics[width=0.45\textwidth]{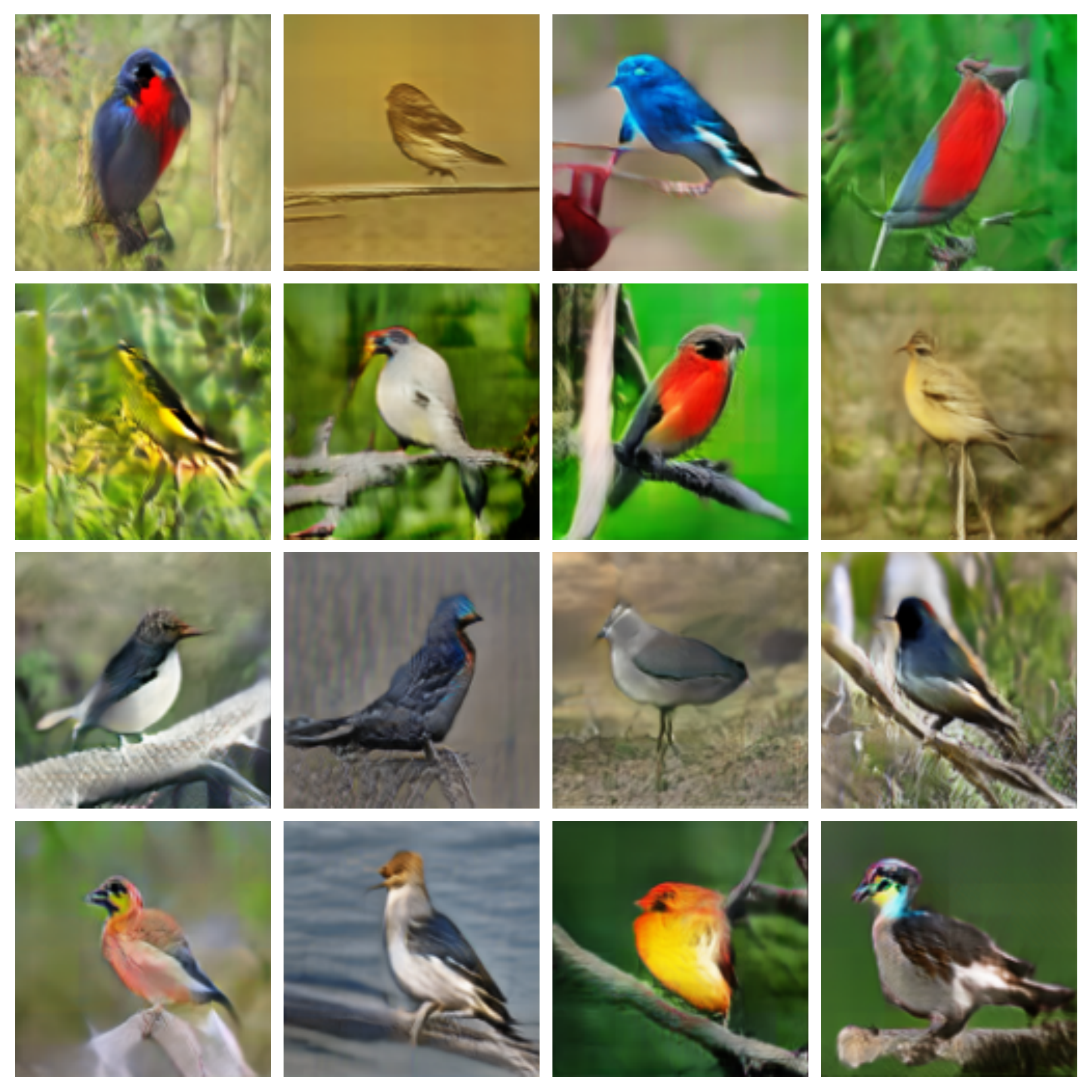}
    \end{subfigure} \hfill
    \begin{subfigure}%{0.48\textwidth}
        \centering
        \includegraphics[width=0.45\textwidth]{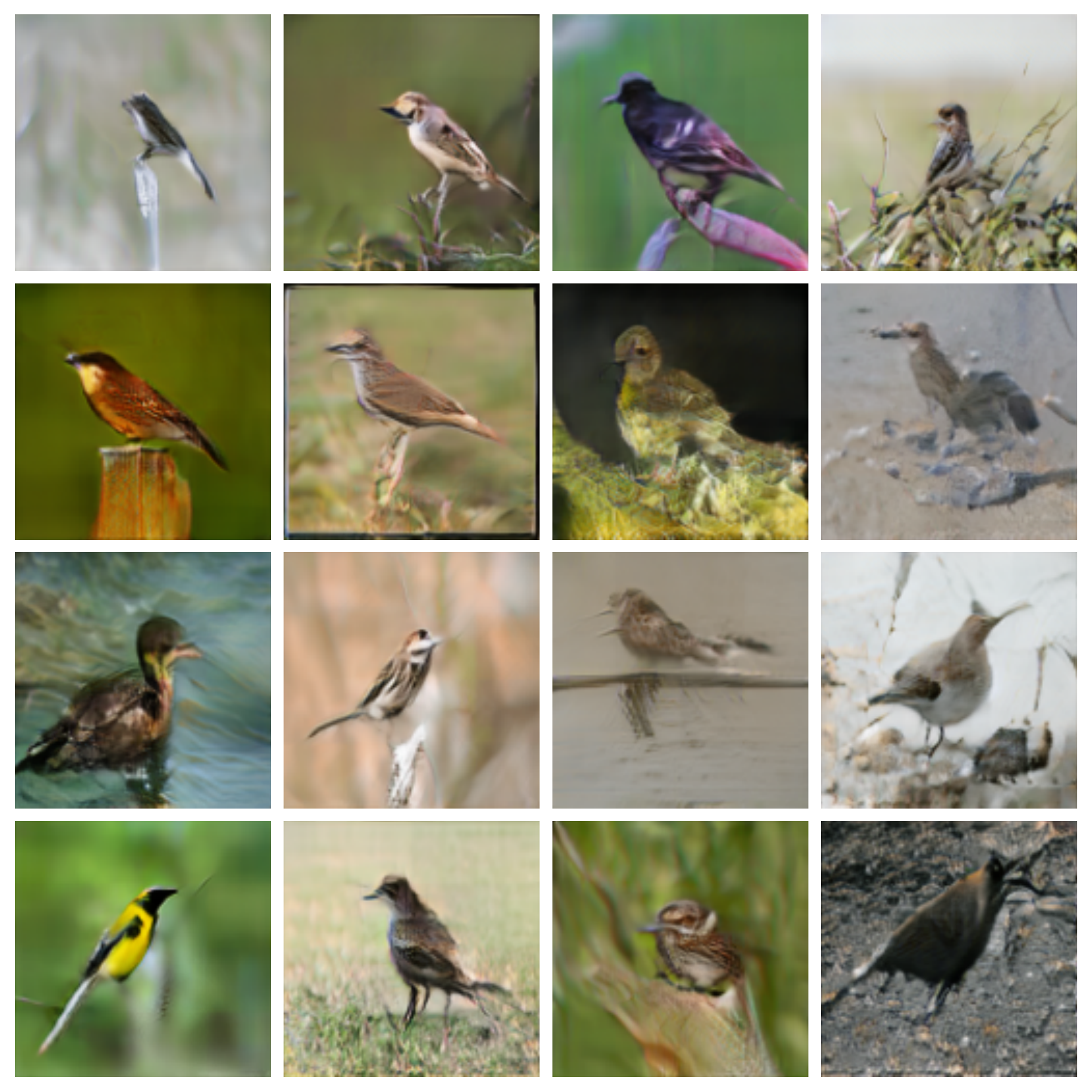}
    \end{subfigure}
    % \vspace{-3mm}
    \caption{\textbf{Sample Diversity on CUB200:} We showcase samples from a BigGAN pretrained on imagenet \& finetuned on CUB200 with the AMAT procedure (left) and from an identical BigGAN finetuned without AMAT (right). Observe that while the sample quality is good for both, the samples generated with AMAT are more colorful \& diverse, with bright reds and yellow against several backgrounds. While the samples from vanilla fine-tuning are restricted to whites/grays \& only a hint of color.  
    }
    \label{fig:cub200}
    \vspace{-6mm}
\end{figure*}

We test our proposed method on several synthetic and real datasets \& report a consistent increase in performance on GAN evaluation metrics such as
Inception Score \citep{salimans2016improved} and Frech\'et Inception Distance \citep{heusel2017gans} with our proposed AMAT. We also showcase our performance in the GAN fine-tuning regime with samples on the CUB200 dataset \citep{WelinderEtal2010} which qualitatively are more colorful and diverse than an identical BigGAN without AMAT (Figure \ref{fig:cub200}). 

\begin{table*}[t]

\resizebox{\textwidth}{!}{% use resizebox with textwidth
\begin{tabular}{ccccc|c c}
\toprule
                              & GAN         & UnrolledGAN         & D2GAN               & RegGAN              & DCGAN & with AMAT \\ \midrule
\# Modes covered       & $628.0 \pm 140.9$   & $817.4 \pm 37.9$    & $1000 \pm 0.0$     & $ 955.5\pm18.7$     & $849.6\pm 62.7$    & $\mathbf{1000 \pm 0.0}$             \\
KL (samples $\Vert$ data) & $2.58\pm 0.75$  & $1.43 \pm 0.12$     & $0.080 \pm 0.01$                 & $0.64 \pm 0.05$     & $0.73\pm0.09$     &     $0.078\pm0.01$      \\ \bottomrule

\end{tabular}
}
\vspace{3mm}
\caption{\textbf{Quantitative Results on the Stacked MNIST dataset}: Applying our proposed adaptive multi adversarial training (AMAT) procedure to a simple DCGAN achieves perfect mode coverage, better than many existing methods for mode collapse.}
% \vspace{-4mm}
\label{tab:stacked}
\end{table*}

\begin{table*}[t]
    \smaller
    \resizebox{\textwidth}{!}{
    \begin{tabular}{c| cc|cc|cc}
        \toprule
               &                      &                        &  GAN-NS               &   AMAT +             &                  & AMAT +               \\
        Model  &  D2GAN               & MicrobatchGAN          &  w/ ResNet            &   GAN-NS             & DCGAN            & DCGAN          \\ \midrule
        IS     & $7.15 \pm 0.07$      & $6.77$                 &  $6.7 \pm 0.06$                & $\mathbf{8.1 \pm 0.04}$       &  $6.03\pm0.05$   & $\mathbf{6.32\pm0.06}$ \\
        FID    &  -                   & -                      &  $28.91$              & $\mathbf{16.35}$     &  $33.42$         & $\mathbf{30.14}$ \\ \midrule
               & WGAN-GP              & AMAT +                 &                       &AMAT +                &                  & AMAT +         \\
        Model  & w/ ResNet            & WGAN-GP                & SN-GAN                &  SN-GAN              & BigGAN           & BigGAN         \\ \midrule
        IS     & $7.59\pm0.10$        & $\mathbf{7.80\pm0.07}$ &$ 8.22\pm0.05$         &$\mathbf{8.34\pm0.04}$&  $9.22$          & $\mathbf{9.51\pm0.06}$ \\
        FID    & $19.2$            & $\mathbf{17.2}$     &$14.21$             &$\mathbf{13.8}$    &  $8.94$      & $\mathbf{6.11}$ \\
        \bottomrule
    \end{tabular}
    }
    \vspace{3mm}
    \centering
    \caption{\textbf{Quantitative Results on CIFAR10}: We benchmark AMAT against several other multi-adversarial baselines as well as on several GAN architectures across all of which we observe a consistent performance increase.}
    \label{tab:cifar}
\end{table*}
\vspace{-3mm}
\subsection{Synthetic Data}
\vspace{-1mm}
We utilize the proposed synthetic data generation procedure with randomly initialized normalizing flows to visualize the training process of a simple DCGAN \citep{radford2015unsupervised}. Figure \ref{fig:oscillation} visualizes such a training process for a simple bimodal distribution. Observing the pattern of generated samples over the training iteration and the shifting discriminator landscape, we note a clear mode oscillation issue present in the generated samples driven by the shifting discriminator output distribution. Focusing on a single fixed real point in space at any of the modes, we see a clear oscillation in the discriminator output probabilities strongly indicating the presence of catastrophic forgetting in the discriminator network. 
%Further such visualizations on more complex distributions (toy $8$-D Gaussian rings) are added in the Supplementary Material. 
%Appendix~\ref{appendix:synthetic}.  

\noindent \textbf{Effect of Data Complexity on Mode Collapse}: We use the flexibility in choosing transformations $g_i$ to generate datasets of various data distribution complexities as presented in Table $\ref{tab:synthetic}$. Choosing $g(z)$ with successively more complicated transformations can produce synthetic datasets of increasing complexity, the first five of which we roughly classify as {\fontfamily{lmtt}\selectfont Levels}. The {\fontfamily{lmtt}\selectfont Levels} are generated by using simple transforms such as identitym constant mapping, small Multi layer perceptrons and well conditioned linear transforms ($\mathbf{A}$). 

On this benchmark, we investigate mode collapse across different optimizers such as SGD \& ADAM \citep{kingma2014adam} on several popular GAN variants such as the non-saturating GAN Loss (GAN-NS) \citep{goodfellow2014generative}, WGAN \citep{arjovsky2017wasserstein} and also methods targeting mitigating mode collapse specifically such as Unrolled GAN \citep{metz2016unrolled} and D2GAN \citep{nguyen2017dual}. 
We show results of our proposed AMAT training procedure with a simple GAN-NS, which matches performance with other more complicated mode collapse specific GAN architectures, all of which are robust to mode collapse up to {\fontfamily{lmtt}\selectfont Level IV}. 
%The procedure can be extended to even more complex distributions than {\fontfamily{lmtt}\selectfont Level V}, but 
In practice we find all benchmarked methods to collapse at {\fontfamily{lmtt}\selectfont Level V}. Thus, in contrast to other simple datasets like MNIST~\citep{lecun1998mnist}, Gaussian ring, or Stacked MNIST~\citep{metz2016unrolled}, the complexity of our synthetic dataset can be arbitrarily tuned up or down to gain insight into the training and debugging of GAN via visualizations.
\begin{table*}[t]
\vspace{-2mm}
    \centering
    \resizebox{\textwidth}{!}{
    \begin{tabular}{c| ccccc| c}
        \toprule
        Effect     & Large $|\displaystyle \sD|$        &   Spawn too late       &  Greedy $\displaystyle \nabla$D     &  Random for fake              &         $\mathbf{1}$-hot weight  & Proposed \\
        Ablation   &     Small $\alpha$, Short $T_t$    &  Long $T_t$ schedule   & $\epsilon = 0$        &   $\epsilon$-greedy for real  & vector $\vm$ & Method                \\ \midrule
        IS         &  $8.83 \pm 0.04$                   & $9.28 \pm 0.08$        &      $9.31 \pm 0.06$  & $8.95 \pm 0.04$        & $9.25 \pm 0.05$ & $9.51 \pm 0.06$                \\
        FID        &  $14.23$                           &   $9.37$      &  $8.6$              &     $12.5$          & $9.25$   & $6.11$               \\ 
        \bottomrule
    \end{tabular}
    }
    \vspace{3mm}
     \caption{\textbf{BigGAN + AMAT Ablations on CIFAR10} (A) A spawning condition with small $\alpha$ and short warmup schedule that leads to large number of discriminators (>$7$) (B) Long warm-up schedules that spawn new discriminators late into training (C) A greedy strategy for assigning responsibility of fake samples ($\epsilon = 0$) (D) Flipping the data splitting logic with responsibilities of fake samples being random and of real being $\epsilon$-greedy (E) Choosing  the discriminator with lowest score for updating Generator instead of soft random weighting.}
     \vspace{-2mm}
    \label{tab:ablations}
\end{table*}

\begin{figure*}
    \centering
    \begin{subfigure}%{0.48\textwidth}
        \centering
        \includegraphics[width=0.48\textwidth]{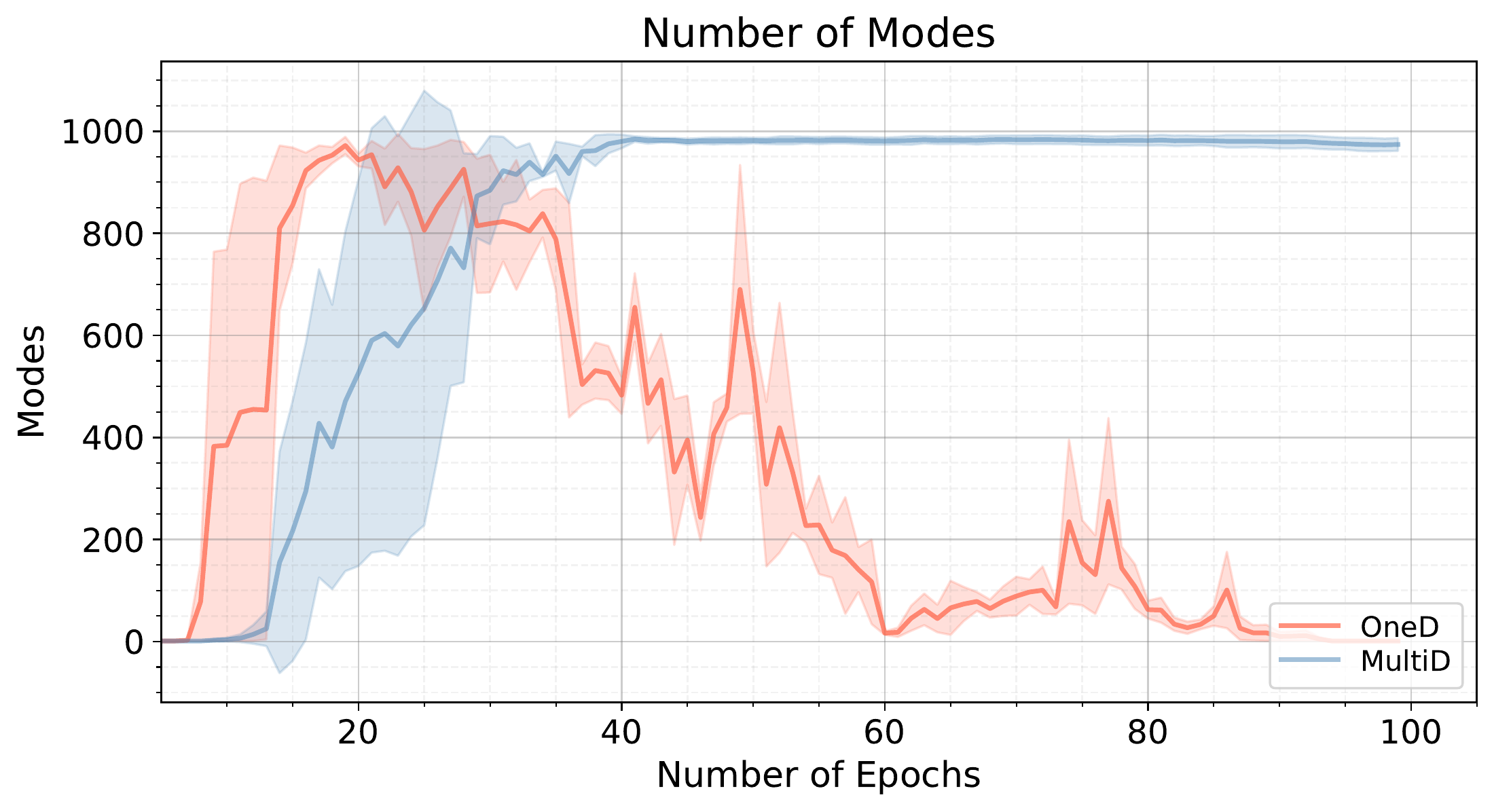}
    \end{subfigure}
    \begin{subfigure}%{0.48\textwidth}
        \centering
        \includegraphics[width=0.48\textwidth]{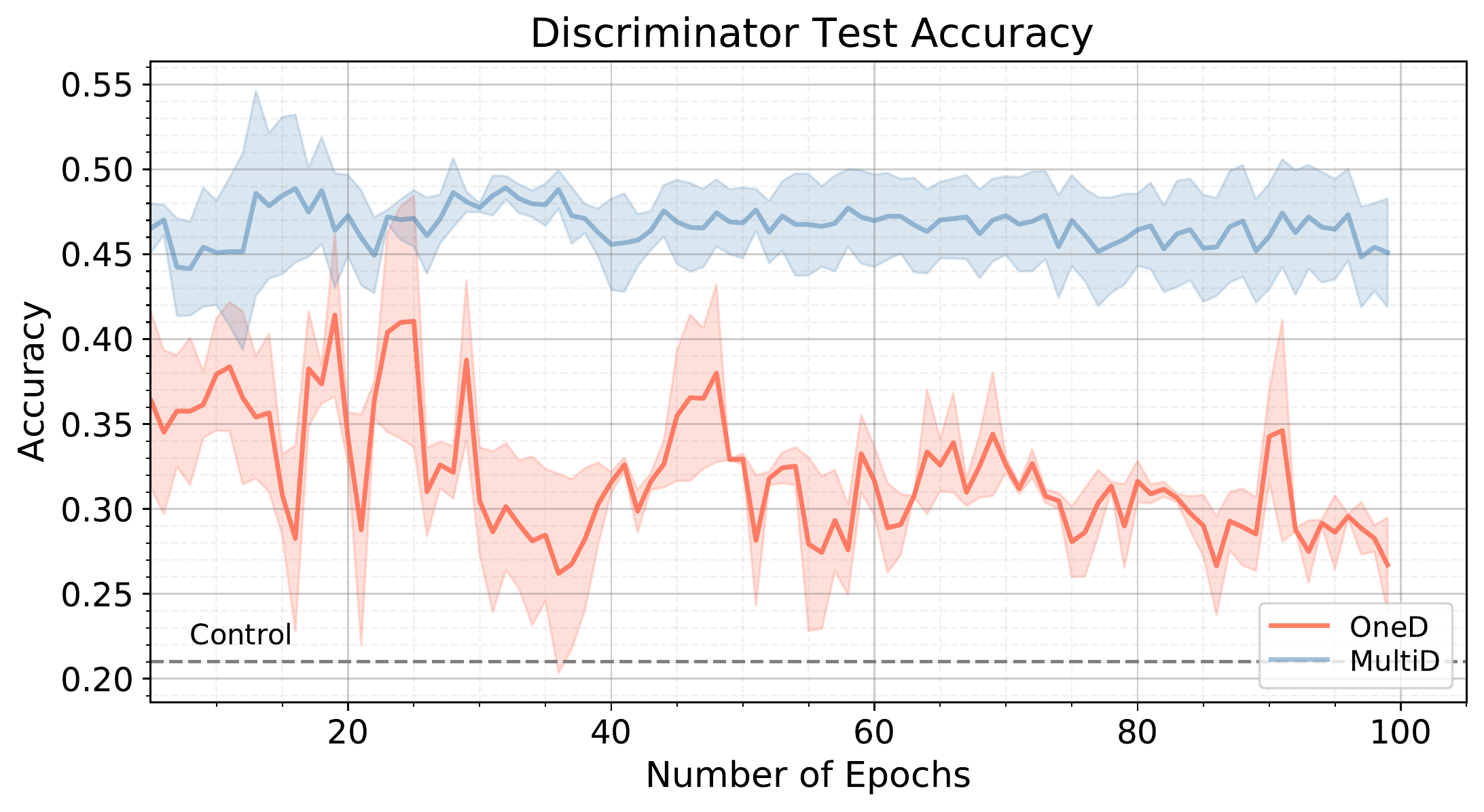}
    \end{subfigure}
    \vspace{1mm}
    \caption{{\textbf{Investigating the forgetting-collapse interplay:} We investigate our hypothesis that catastrophic forgetting is associated with mode collapse. On the left pane, we plot the magnitude of mode collapse by counting the number of modes produced by the generator. On the right pane, we assess the quality of the discriminator features by plotting the accuracy of linear classifier on top of the discriminator features at each epoch. In the original model, the coverage of modes and the quality of discriminator features are both low and decreasing. In particular, the test accuracy from the discriminator's features drops almost to randomly initialized weights (shown as \textit{control}). On the other hand, adding AMAT (\textit{MultiD}) dramatically improves both mode coverage and the discriminator test accuracy.
    }}
    \label{fig:classification}
    \vspace{-4mm}
\end{figure*}
\vspace{-3mm}
\subsection{Stacked MNIST} 
\vspace{-2mm}
We also benchmark several models on the Stacked MNIST dataset following \citep{metz2016unrolled, srivastava2017veegan}. Stacked MNIST is an extension of the popular MNIST dataset~\citep{lecun1998gradient} where each image is expanded in the channel dimension to $28 \times 28 \times 3$ by concatenating $3$ single channel images. The resulting dataset has a $1000$ overall modes. We measure the number of modes covered by the generator as the number of classes that are generated at least once within a pool of $25,600$ sampled images. The class of the generated sample is identified with a pretrained MNIST classifier operating channel wise on the original stacked MNIST image. 
% We also measure the KL divergence between the label distribution predicted by the MNIST classifier in the previous experiment and the expected data distribution.
\newline
\newline
\noindent \textbf{Understanding the forgetting-collapse interplay}: In Section \ref{sec:intro}, we discuss our motivation for studying catastrophic forgetting for mitigating mode collapse. We also design an investigative experiment to explicitly observe this interplay by comparing the number of modes the generator learns against the quality of features the discriminator learns throughout GAN training on the stacked MNIST dataset. We measure the number of modes captured by the generator through a pre-trained classification network trained in a supervised learning fashion and frozen throughout GAN training. To measure the amount of \emph{`forgetting`} in discriminator, we extract features of real samples from the penultimate layer of the discriminator and train a small classifier on the real features for detecting real data mode. This implicitly indicates the quality and information contained in the the discriminator extracted features. However, the performance of classification network on top of discriminator features is confounded by the capacity of the classification network itself. Hence we do a control experiment, where we train the same classifier on features extracted from a randomly initialized discriminator, hence fixing a lower-bound to the classifier accuracy.
\begin{table*}[t]
\centering

\resizebox{\textwidth}{!}{
\begin{tabular}{c|c|c|c|c|c|c|c|c|c|c|c}
\toprule
Classes  & Plane               & Car               & Bird               & Cat               & Deer           & Dog               & Frog               & Horse               & Ship               & Truck & Avg              \\ \midrule
 BigGAN     & $24.23$         & $12.32$         & $24.85$         & $21.21$         & $12.81$        & $22.74$         & $17.95$         & $13.16$         & $12.11$         & $18.39$ & $8.94$         \\
                     $+$ AMAT & $20.50$         & $10.30$         & $23.48$         & $18.48$         & $11.51$       & $19.41$         & $11.50$         & $12.24$         & $10.69$         & $12.94$  & $6.11$       \\ 
                      $\Delta \%$    & $18.2$           &  $19.6$               &       $5.8$          &     $14.8$            &   $11.3$        &     $17.2$            & $\mathbf{56.1}$                &         $7.5$        &      $11.7$           &     $\mathbf{42.1}$            &  $\mathbf{46.3}$    
                      \\ \bottomrule
\end{tabular}
}
\vspace{3mm}
\caption{\textbf{Per-class FID on CIFAR10}: FID improves consistently across all classes.}
\label{tab:classwise}
\vspace{-3mm}
\end{table*}
Referring to Figure \ref{fig:classification}, we observe a clear relation between the number of modes the generator covers at an iteration and the accuracy of the classification network trained on the discriminator features at the same iteration. In the vanilla single discriminator scenario, the classification accuracy drops significantly, indicating a direct degradation of the discriminative features which is followed by a complete collapse of G. In the collapse phase, the discriminator's learnt features are close to random with the classification accuracy being close to that of the control experiment. This indicates the presence of significant catastrophic forgetting in the the discriminator network. 

In contrast, training the same generator with the proposed AMAT procedure leads to stable training with almost all the modes being covered. The classification accuracy increasing before saturation. Catastrophic forgetting is \textit{effectively sidestepped} by adaptive multi adversarial training which produces stable discriminative features during training that provide a consistent training signal to the generator thereby covering all the modes with little degradation.
\vspace{-1mm}
\subsection{CIFAR10}
\vspace{-1mm}
We extensively benchmark AMAT on several GAN variants including unconditional methods such as DCGAN \citep{radford2015unsupervised}, ResNet-WGAN-GP \citep{gulrajani2017improved,he2016deep} \& SNGAN \citep{miyato2018spectral} and also conditional models such as BigGAN \citep{brock2018large}. Table \ref{tab:cifar} shows the performance gain on standard GAN evaluation metrics such as Inception Score and Fr\'echet distance of several architectures when trained with AMAT procedure. The performance gains indicate effective curbing of catastrophic forgetting in the discriminator with multi adversarial training. We use the public evaluation code from SNGAN \citep{miyato2018spectral} for evaluation. Despite having components such as spectral normalization, diversity promoting loss functions, additional R1 losses \& other stable training tricks that might affect catastrophic forgetting to different extents, we observe a consistent increase in performance across all models. Notably the ResNet GAN benefits greatly with AMAT despite a powerful backbone -- with IS improving from $6.7$ to $8.1$, indicating that the mode oscillation problem is not mitigated by simply using a better model. 

AMAT improves performance by over $35\%$ even on a well performing baseline such as BigGAN (Table \ref{tab:cifar}). We investigate classwise FID scores of a vanilla BigGAN and an identical BigGAN + AMAT on CIFAR10 and report the results in Table $\ref{tab:classwise}$. Performance improves across all classes with previously poor performing classes such as `Frog'  \& `Truck' experiencing the most gains. Further, we ablate several key components of AMAT procedure on the BigGAN architecture with results reported in Table \ref{tab:ablations}. We observe all elements to be critical to overall performance. Specifically, having a moderate $\alpha$ schedule to avoid adding too many discriminators is critical. Another viable design choice is to effectively flip the algorithm's logic and instead choose the fake points randomly while being $\epsilon$ greedy on the real points. This strategy performs well on simple datasets but loses performance with BigGAN on CIFAR10 (Table \ref{tab:ablations}). In all experiments, the computational time during inference is the same as the base model, \textit{irrespective of the number of discriminators} added during the training, since only a single generator is trained with AMAT and all the discriminators are discarded.
\vspace{-4mm}
\section{Conclusion}
\vspace{-3mm}
In summary, motivated from the observation of catastrophic forgetting in the discriminator, we propose a new adaptive GAN training framework that adds additional discriminators to prevent mode collapse. We show that our method can be added to existing GAN frameworks to prevent mode collapse, generate more diverse samples and improve FID \& IS. In future, we plan to apply AMAT to fight mode collapse in high resolution image generation settings. 
\vspace{-3mm}
\subsubsection*{Acknowledgements}
\vspace{-2mm}
We thank Jathushan Rajasegaran for his helping with the forgetting-collapse interplay experiments and Taesung Park for feedback and comments on the early drafts of this paper.

% \bibliography{egbib}

\clearpage

\section*{Appendix}

\section*{CIFAR10 Experiments}
\label{appendix:cifarexp}

\subsection*{BigGAN + AMAT Experiments}

For the baseline we use the author's official PyTorch implementation \footnote{\href{https://github.com/ajbrock/BigGAN-PyTorch}{https://github.com/ajbrock/BigGAN-PyTorch}}. For our experiments on AMAT $+$ BigGAN, we kept the optimizer as Adam \citep{kingma2014adam} and used the hyperparameters $\beta_1 = 0.0, \beta_2 = 0.9$. We did not change the model architecture parameters in any way. The best performance was achieved with learning rate $=0.0002$ for both the Generator and all the Discriminators. The batch size for the $G$ and all $D$ is $50$, and the latent dimension is chosen as $128$. The initial value of $T_t = 5$ epochs, and after the first discriminator is added, $T_t$ is increased by 5 epochs every $T_t$ epochs. The initial value of $\alpha_t = 1.5$, and it is increased by a factor of $3.5$ every time a discriminator is added. To check whether to add another discriminator or not, we use 10 exemplar images, 1 from each CIFAR10 class. While assigning datapoints to each discriminator, we use an epsilon greedy approach. We chose $\epsilon = 0.25$, where the datapoint is assigned to a random discriminator with a probability $\epsilon$. The number of discriminator(s) updates per generator update is fixed at $4$. We also use exponential moving average for the generator weights with a decay of $0.9999$. 

\subsection*{SN-GAN + AMAT Experiments}

We used the SN-GAN implementation from \href{https://github.com/GongXinyuu/sngan.pytorch}{https://github.com/GongXinyuu/sngan.pytorch}, which is the PyTorch version of the authors' Chainer implementation \href{https://github.com/pfnet-research/sngan_projection}{https://github.com/pfnet-research/sngan\_projection}. We kept the optimiser as Adam and used the hyperparameters $\beta_1 = 0.0, \beta_2 = 0.9$. The batch size for generator is $128$ and for the discriminators is $64$, and the latent dimension is $128$. The initial learning rate is $0.0002$ for both generator and the discriminators. The number of discriminator(s) updates per generator update is fixed at $7$. The initial value of $T_t = 2$ epochs, and is increased by $1$ epoch after every discriminator is added. $\alpha_t$ is initialized as $1.5$, and is increased by a factor of $1.3$ after a discriminator is added, till $20$ epochs, after which it is increased by a factor of $3.0$. These larger increases in $\alpha_t$ are required to prevent too many discriminators from being added over all iterations. We chose $\epsilon = 0.3$, where the datapoint is assigned to a random discriminator with a probability $\epsilon$.  We use 10 exemplar images, 1 from each CIFAR10 class.

\textbf{ResNet GAN:} We use the same ResNet architecture as above, but remove the spectral normalization from the model. The optimizer parameters, learning rate and batch sizes remain the same as well. The number of discriminator(s) updates per generator update is fixed at $5$. The initial value of $T_t = 10$ epochs, and is increased by $5$ epochs after every discriminator is added. $\alpha_t$ is initialized as $1.5$, and is increased by a factor of $2.0$ after a discriminator is added. We chose $\epsilon = 0.2$, where the datapoint is assigned to a random discriminator with a probability $\epsilon$.  We use 10 exemplar images, 1 from each CIFAR10 class.

\textbf{ResNet WGAN-GP:} In the above model, the hinge loss is replaced by the Wasserstein loss with gradient penalty. The optimizer parameters, learning rate and batch sizes remain the same as well.The number of discriminator(s) updates per generator update is fixed at $2$. The initial value of $T_t = 5$ epochs, and is increased by $5$ epochs after a discriminator is added. $\alpha_t$ is initialized as $1.5$, and is increased by a factor of $3.0$ after a discriminator is added. We chose $\epsilon = 0.2$, where the datapoint is assigned to a random discriminator with a probability $\epsilon$. We use 10 exemplar images, 1 from each CIFAR10 class. These images are chosen randomly from each class, and may not be the same as the ones for other CIFAR10 experiments.

\subsection*{DCGAN + AMAT Experiments}

We used standard CNN models for our DCGAN as shown in Table \ref{tab:dcganArch}. We use Adam optimizer with hyperparameters $\beta_1 = 0.0, \beta_2 = 0.9$. The learning rate for generator was $0.0002$, and the learning rate for the discriminator(s) was $0.0001$. The number of discriminators updates per generator was fixed at 1. The initial value of $T_t = 4$ epochs, and is increased $5$ epochs after a discriminator is added. $\alpha_t $ is initialized as $1.5$ and is increased by a factor of $1.5$ every time a discriminator is added. We chose $\epsilon = 0.3$, where the datapoint is assigned to a random discriminator with a probability $\epsilon$. We use 10 exemplar images, 1 from each CIFAR10 class.

\section*{Stacked MNIST Experiments}
Stacked MNIST provides us a test-bed to measure mode collapse. A three channel image is generated by stacking randomly sampled MNIST classes, thus creating a data distribution if 1000 modes. We use this dataset to show that, when generator oscillates to a different set of modes, catastrophic forgetting is induced in discriminator and this prevents the generator to recover previous modes. To study this phenomenon, we need to measure the correlation between number of modes the generator covered and the catastrophic forgetting in the discriminator. Measuring the number of modes is straight forward, we can by simply classify each channels of the generated images using a MNIST pretrained classifier to find its corresponding mode. However, to measure catastrophic forgetting in the discriminator, we use a proxy setting, where we take the high-level features of the real images from the discriminator and train a simple classifier on top of that. The discriminative quality of the features taken from the discriminator indirectly measure the ability of the network to remember the modes. Finally, as a control experiment we randomize the weights of the discriminator, and train a classifiers on the feature taken from randomized discriminator. This is to show that, extra parameters in the classifier does not interfere our proxy measure for the catastrophic forgetting. Finally, we train a DCGAN with a single discriminator, and a similar DCGAN architecture with our proposed AMAT procedure, and measure the number of modes covered by the generator and the accuracy of the discriminator.

\section*{A Fair comparison on discriminator capacity}

Our AMAT approach incrementally adds new discriminators to the GAN frameworks, and its overall capacity increases over time. Therefore, it is not fair to compare a model with AMAT training procedure with its corresponding the single discriminator model. As a fair comparison to our AMAT algorithm, we ran single discriminator model with approximately matching its discriminator capacity to the final AMAT model. For example, SN-GAN with AMAT learning scheme uses 4 discriminators at the end of its training. Therefore we use a  discriminator with four times more parameters for the single discriminator SN-GAN model. This is done by increasing the convolutional fillters in the discriminator. Table~\ref{tab:fairD} shows that, even after matching the network capacity, the single discriminator models do not perform well as compared to our AMAT learning.

\begin{table*}[!t]
    \centering
    \resizebox{\textwidth}{!}{
    \begin{tabular}{c|c| ccccc}
        \toprule
        &   Scores     &  DCGAN   &   ResNetGAN     &  WGAN-GP   &  SN-GAN   &   BigGAN  \\ \midrule
      &   \#of Param of D &    1.10 M      &       3.22 M          &    2.06 M       &    4.20 M       &    8.42 M     \\ 
  w/o AMAT      &   IS         &  5.97 $\pm$ 0.08     &  6.59 $\pm$ 0.09  &       7.72 $\pm$ 0.06          &    8.24 $\pm$ 0.05 &     9.14 $\pm$ 0.05      \\ 
        &   FID        &   34.7       &      36.4           &     19.1      &     14.5      &     10.5                 \\  \midrule
        &   \#of Param of D &   1.02 M       &        3 $\times$ 1.05 M         &    2 $\times$ 1.05 M       &    4 $\times$ 1.05 M       &    8.50 M      \\ 
+ AMAT   &   IS         &   6.32 $\pm$ 0.06       &        8.1 $\pm$ 0.04         &     7.80 $\pm$ 0.07      &    8.34 $\pm$ 0.04      &     9.51 $\pm$ 0.06      \\ 
        &   FID        &     30.14     &        16.35         &     17.2      &       13.8    &      6.11     \\  \midrule
        \bottomrule
    \end{tabular}
    }
    \vspace{2mm}
    \caption{Increasing network capacity alone does not capture more modes. Even after the discriminator capacity is matched, single discriminator GANs do not perform as well as multi-adversarial GANs with AMAT learning}
    \label{tab:fairD}
\end{table*}

\section*{Synthetic Data Experiments}
\label{appendix:synthetic}

We add flow-based non-linearity (Algorithm \ref{algo:synthetic}) to a synthetic 8-Gaussian ring dataset. We chose $K=5$ as our non-linearity depth and chose a randomly initiated 5 layer MLP as our non-linear functions. We use an MLP as our GAN generator and discriminator (Table \ref{tab:mlpArch}). We use the Adam optimizer with hyperparameters $\beta_1 = 0.0, \beta_2 = 0.9$. The learning rate for the generator and discriminator was $0.0002$. The number of discriminator updates per generator update is fixed to 1, and the batch size is kept 64. The initial value of $T_t = 5$ epochs, and is increased by $10$ every time a discriminator is added. $\alpha_t$ is initialized as $1.5$, and is increased by a factor of $1.5$ after a discriminator is added for the first 50 epochs. After that $\alpha_t$ is increased by a factor of $3$. We chose $\epsilon = 0.25$, where the datapoint is assigned to a random discriminator with a probability $\epsilon$. 1 random datapoint from each of the 8 modes is selected as the exemplar image. 

\begin{figure*}
    % \vspace{-10mm}
    \begin{subfigure}%{0.24\textwidth}
        \centering
        \animategraphics[loop,autoplay,width=0.24\textwidth]{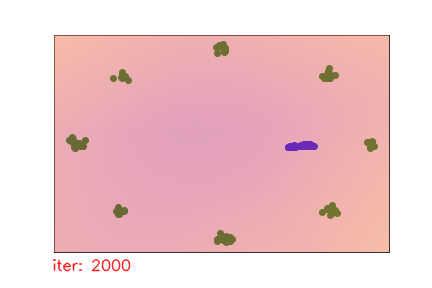}{figs/gif0/}{0}{47}
    \end{subfigure} 
    \begin{subfigure}%{0.24\textwidth}
        \centering
        \animategraphics[loop,autoplay,width=0.24\textwidth]{5}{figs/gen/frame_}{1}{48}
    \end{subfigure} 
    \begin{subfigure}%{0.24\textwidth}
        \centering
        \animategraphics[loop,autoplay,width=0.24\textwidth]{5}{figs/d0/frame_}{1}{48}
    \end{subfigure}
    \begin{subfigure}%{0.24\textwidth}
        \centering
        \animategraphics[loop,autoplay,width=0.24\textwidth]{5}{figs/d1/frame_}{1}{48}
    \end{subfigure}
    % \vspace{-3mm}
    \caption{ \small{\textbf{GAN training visualization}: (Figure contains animated graphics, better viewed in Adobe Acrobat Reader) Training trajectories of an MLP in table \ref{tab:mlpArch} (leftmost panel) and an MLP trained with our AMAT procedure (Algorithm \ref{algo:multi}) (rest three panels) on a $784$-dimensional synthetic dataset. Green dots represent real samples and the blue dots represent the generated samples. The vanilla GAN samples are overlayed against discriminator's output heatmap where the warm yellow color indicates a high probability of being real and cold violet indicates fake. In the AMAT + GAN panels, the discriminator landscapes are shown separately for both discriminators with the second discriminator being spawned at iteration $4000$ (Algorithm \ref{algo:spawn}). The $2$D visualizations of the $784$D data space is facilitated by our synthetic data generation procedure (Algorithm \ref{algo:synthetic}). }}
    \label{fig:eightgaussian}
    % \vspace{-6mm}
\end{figure*}

Figure \ref{fig:eightgaussian} shows the difference in performance of a standard MLP GAN (\ref{tab:mlpArch} and the same MLP GAN with AMAT. The GIF on the lefts shows a cyclic mode collapse due the discriminator suffering from catastrophic forgetting. The same GAN with is able to completely mitigate catastrophic forgetting with just 2 discriminators added during training, on a 728-dimensional synthetic data. 

\begin{table}[!htb]
    % \centering
    % \begin{subtable}%[width=0.45\textwidth]
    \parbox{0.45\linewidth}{\centering
        \begin{tabular}{c}
          \toprule
          \midrule
          $z \in \mathbb{R}^{128} \sim \mathcal{N}(0,I) $\\
          \midrule
          dense $\rightarrow 4 \times 4 \times 512$\\
          \midrule
          $4\times4$, stride$=$2 deconv. BN 256 ReLU\\
          \midrule
          $4\times4$, stride$=$2 deconv. BN 128 ReLU\\
          \midrule
          $4\times4$, stride$=$2 deconv. BN 64 ReLU\\
          \midrule
          $3\times3$, stride$=$1 conv. 3 Tanh\\
          \midrule
          \bottomrule
        \end{tabular}
        \vspace{3mm}
        \caption*{Generator}
        }
    \vspace{1mm}
    % \end{subtable}%
    % \begin{subtable}%{0.45\textwidth}
    \parbox{0.45\linewidth}{\centering
    \begin{tabular}{c}
          \toprule
          \midrule
          $x \in \mathbb{R}^{32 \times 32 \times 3}$\\
          \midrule
          $3\times3$, stride$=$1 conv. 64 lReLU\\
          $4\times4$, stride$=$2 conv. 64 lReLU\\
          \midrule
          $3\times3$, stride$=$1 conv. 128 lReLU\\
          $4\times4$, stride$=$2 conv. 128 lReLU\\
          \midrule
          $3\times3$, stride$=$1 conv. 256 lReLU\\
          $4\times4$, stride$=$2 conv. 256 lReLU\\
          \midrule
          $3\times3$, stride$=$1 conv. 512 lReLU\\
          \midrule
          dense $\rightarrow 1$\\
          \midrule
          \bottomrule
    \end{tabular}
    \vspace{3mm}
    \caption*{Discriminator}
    }
    % \end{subtable}
    \vspace{3mm}
    \caption{DCGAN Architecture for CIFAR10}

    \label{tab:dcganArch}
\end{table}

\begin{table*}[!t]
    \centering
    % \begin{subtable}
    \parbox{0.45\linewidth}{\centering
    \begin{tabular}{c}
          \toprule
          \midrule
          $z \in \mathbb{R}^{25} \sim \mathcal{N}(0,I) $\\
          \midrule
          dense $\rightarrow 128$, BN 128 ReLU\\
          \midrule
          dense $\rightarrow 128$, BN 128 ReLU\\
          \midrule
          dense $\rightarrow 512$, BN 512 ReLU\\
          \midrule
          dense $\rightarrow 1024$, BN 1024 ReLU\\
          \midrule
          dense $\rightarrow 2$, Tanh\\
          \midrule
          \bottomrule
    \end{tabular}
    \vspace{2mm}
    \caption*{Generator}
    }
    % \end{subtable}
    % \begin{subtable}
    \parbox{0.45\linewidth}{\centering
    \begin{tabular}{c}
          \toprule
          \midrule
          $x \in \mathbb{R}^{2}$\\
          \midrule
          dense $\rightarrow 128$ ReLU\\
          \midrule
          dense $\rightarrow 512$ ReLU\\
          \midrule
          dense $\rightarrow 1$ Sigmoid\\
          \midrule
          \bottomrule
    \end{tabular}
    \vspace{2mm}
    \caption*{Discriminator}
    }
    % \end{subtable}
    \vspace{1mm}
    \caption{MLP architecture for Synthetic Dataset}

    \label{tab:mlpArch}
\end{table*}

\clearpage
% % \nocite{langley00}
\bibliography{egbib}

\begin{thebibliography}{47}
\providecommand{\natexlab}[1]{#1}
\providecommand{\url}[1]{\texttt{#1}}
\expandafter\ifx\csname urlstyle\endcsname\relax
  \providecommand{\doi}[1]{doi: #1}\else
  \providecommand{\doi}{doi: \begingroup \urlstyle{rm}\Url}\fi

\bibitem[Arjovsky et~al.(2017)Arjovsky, Chintala, and
  Bottou]{arjovsky2017wasserstein}
Martin Arjovsky, Soumith Chintala, and L{\'e}on Bottou.
\newblock Wasserstein gan.
\newblock \emph{arXiv preprint arXiv:1701.07875}, 2017.

\bibitem[Bau et~al.(2019)Bau, Zhu, Wulff, Peebles, Strobelt, Zhou, and
  Torralba]{bau2019seeing}
David Bau, Jun-Yan Zhu, Jonas Wulff, William Peebles, Hendrik Strobelt, Bolei
  Zhou, and Antonio Torralba.
\newblock Seeing what a gan cannot generate.
\newblock In \emph{Proceedings of the International Conference Computer Vision
  (ICCV)}, 2019.

\bibitem[Bousmalis et~al.(2018)Bousmalis, Irpan, Wohlhart, Bai, Kelcey,
  Kalakrishnan, Downs, Ibarz, Pastor, Konolige, et~al.]{bousmalis2018using}
Konstantinos Bousmalis, Alex Irpan, Paul Wohlhart, Yunfei Bai, Matthew Kelcey,
  Mrinal Kalakrishnan, Laura Downs, Julian Ibarz, Peter Pastor, Kurt Konolige,
  et~al.
\newblock Using simulation and domain adaptation to improve efficiency of deep
  robotic grasping.
\newblock In \emph{2018 IEEE international conference on robotics and
  automation (ICRA)}, pages 4243--4250. IEEE, 2018.

\bibitem[Brock et~al.(2018)Brock, Donahue, and Simonyan]{brock2018large}
Andrew Brock, Jeff Donahue, and Karen Simonyan.
\newblock Large scale gan training for high fidelity natural image synthesis.
\newblock \emph{arXiv preprint arXiv:1809.11096}, 2018.

\bibitem[Chen et~al.(2018)Chen, Zhai, and Houlsby]{chen2018self}
Ting Chen, Xiaohua Zhai, and Neil Houlsby.
\newblock Self-supervised gan to counter forgetting.
\newblock \emph{arXiv preprint arXiv:1810.11598}, 2018.

\bibitem[Durugkar et~al.(2016)Durugkar, Gemp, and
  Mahadevan]{durugkar2016generative}
Ishan Durugkar, Ian Gemp, and Sridhar Mahadevan.
\newblock Generative multi-adversarial networks.
\newblock \emph{arXiv preprint arXiv:1611.01673}, 2016.

\bibitem[Fernando et~al.(2017)Fernando, Banarse, Blundell, Zwols, Ha, Rusu,
  Pritzel, and Wierstra]{fernando2017pathnet}
Chrisantha Fernando, Dylan Banarse, Charles Blundell, Yori Zwols, David Ha,
  Andrei~A Rusu, Alexander Pritzel, and Daan Wierstra.
\newblock Pathnet: Evolution channels gradient descent in super neural
  networks.
\newblock \emph{arXiv preprint arXiv:1701.08734}, 2017.

\bibitem[Goodfellow et~al.(2014)Goodfellow, Pouget-Abadie, Mirza, Xu,
  Warde-Farley, Ozair, Courville, and Bengio]{goodfellow2014generative}
Ian Goodfellow, Jean Pouget-Abadie, Mehdi Mirza, Bing Xu, David Warde-Farley,
  Sherjil Ozair, Aaron Courville, and Yoshua Bengio.
\newblock Generative adversarial nets.
\newblock In \emph{Advances in neural information processing systems}, pages
  2672--2680, 2014.

\bibitem[Gulrajani et~al.(2017)Gulrajani, Ahmed, Arjovsky, Dumoulin, and
  Courville]{gulrajani2017improved}
Ishaan Gulrajani, Faruk Ahmed, Martin Arjovsky, Vincent Dumoulin, and Aaron~C
  Courville.
\newblock Improved training of wasserstein gans.
\newblock In \emph{Advances in neural information processing systems}, pages
  5767--5777, 2017.

\bibitem[He et~al.(2016)He, Zhang, Ren, and Sun]{he2016deep}
Kaiming He, Xiangyu Zhang, Shaoqing Ren, and Jian Sun.
\newblock Deep residual learning for image recognition.
\newblock In \emph{Proceedings of the IEEE conference on computer vision and
  pattern recognition}, pages 770--778, 2016.

\bibitem[Heusel et~al.(2017)Heusel, Ramsauer, Unterthiner, Nessler, and
  Hochreiter]{heusel2017gans}
Martin Heusel, Hubert Ramsauer, Thomas Unterthiner, Bernhard Nessler, and Sepp
  Hochreiter.
\newblock Gans trained by a two time-scale update rule converge to a local nash
  equilibrium.
\newblock In \emph{Advances in neural information processing systems}, pages
  6626--6637, 2017.

\bibitem[Hoang et~al.(2018)Hoang, Nguyen, Le, and Phung]{hoang2018mgan}
Quan Hoang, Tu~Dinh Nguyen, Trung Le, and Dinh Phung.
\newblock Mgan: Training generative adversarial nets with multiple generators.
\newblock In \emph{International Conference on Learning Representations}, 2018.

\bibitem[Hoffman et~al.(2018)Hoffman, Tzeng, Park, Zhu, Isola, Saenko, Efros,
  and Darrell]{hoffman2018cycada}
Judy Hoffman, Eric Tzeng, Taesung Park, Jun-Yan Zhu, Phillip Isola, Kate
  Saenko, Alexei Efros, and Trevor Darrell.
\newblock Cycada: Cycle-consistent adversarial domain adaptation.
\newblock In \emph{International conference on machine learning}, pages
  1989--1998. PMLR, 2018.

\bibitem[Karami et~al.(2019)Karami, Schuurmans, Sohl-Dickstein, Dinh, and
  Duckworth]{karami2019invertible}
Mahdi Karami, Dale Schuurmans, Jascha Sohl-Dickstein, Laurent Dinh, and Daniel
  Duckworth.
\newblock Invertible convolutional flow.
\newblock In \emph{Advances in Neural Information Processing Systems}, pages
  5635--5645, 2019.

\bibitem[Karras et~al.(2019)Karras, Laine, and Aila]{karras2019style}
Tero Karras, Samuli Laine, and Timo Aila.
\newblock A style-based generator architecture for generative adversarial
  networks.
\newblock 2019.

\bibitem[Karras et~al.(2020)Karras, Laine, Aittala, Hellsten, Lehtinen, and
  Aila]{karras2020analyzing}
Tero Karras, Samuli Laine, Miika Aittala, Janne Hellsten, Jaakko Lehtinen, and
  Timo Aila.
\newblock Analyzing and improving the image quality of stylegan.
\newblock 2020.

\bibitem[Kingma and Ba(2014)]{kingma2014adam}
Diederik~P Kingma and Jimmy Ba.
\newblock Adam: A method for stochastic optimization.
\newblock \emph{arXiv preprint arXiv:1412.6980}, 2014.

\bibitem[Kirkpatrick et~al.(2017)Kirkpatrick, Pascanu, Rabinowitz, Veness,
  Desjardins, Rusu, Milan, Quan, Ramalho, Grabska-Barwinska,
  et~al.]{kirkpatrick2017overcoming}
James Kirkpatrick, Razvan Pascanu, Neil Rabinowitz, Joel Veness, Guillaume
  Desjardins, Andrei~A Rusu, Kieran Milan, John Quan, Tiago Ramalho, Agnieszka
  Grabska-Barwinska, et~al.
\newblock Overcoming catastrophic forgetting in neural networks.
\newblock \emph{Proceedings of the national academy of sciences}, 114\penalty0
  (13):\penalty0 3521--3526, 2017.

\bibitem[LeCun(1998)]{lecun1998mnist}
Yann LeCun.
\newblock The mnist database of handwritten digits.
\newblock \emph{http://yann. lecun. com/exdb/mnist/}, 1998.

\bibitem[LeCun et~al.(1998)LeCun, Bottou, Bengio, and
  Haffner]{lecun1998gradient}
Yann LeCun, L{\'e}on Bottou, Yoshua Bengio, and Patrick Haffner.
\newblock Gradient-based learning applied to document recognition.
\newblock \emph{Proceedings of the IEEE}, 86\penalty0 (11):\penalty0
  2278--2324, 1998.

\bibitem[Liang et~al.(2018)Liang, Li, Wang, and Carin]{liang2018generative}
Kevin~J Liang, Chunyuan Li, Guoyin Wang, and Lawrence Carin.
\newblock Generative adversarial network training is a continual learning
  problem.
\newblock \emph{arXiv preprint arXiv:1811.11083}, 2018.

\bibitem[Lin et~al.(2018)Lin, Khetan, Fanti, and Oh]{lin2018pacgan}
Zinan Lin, Ashish Khetan, Giulia Fanti, and Sewoong Oh.
\newblock Pacgan: The power of two samples in generative adversarial networks.
\newblock In \emph{Advances in neural information processing systems}, pages
  1498--1507, 2018.

\bibitem[McClelland et~al.(1995)McClelland, McNaughton, and
  O'Reilly]{mcclelland1995there}
James~L McClelland, Bruce~L McNaughton, and Randall~C O'Reilly.
\newblock Why there are complementary learning systems in the hippocampus and
  neocortex: insights from the successes and failures of connectionist models
  of learning and memory.
\newblock \emph{Psychological review}, 102\penalty0 (3):\penalty0 419, 1995.

\bibitem[McCloskey and Cohen(1989)]{mccloskey1989catastrophic}
Michael McCloskey and Neal~J Cohen.
\newblock Catastrophic interference in connectionist networks: The sequential
  learning problem.
\newblock In \emph{Psychology of learning and motivation}, volume~24, pages
  109--165. Elsevier, 1989.

\bibitem[Menon et~al.(2020)Menon, Damian, Hu, Ravi, and Rudin]{Menon_2020_CVPR}
Sachit Menon, Alexandru Damian, Shijia Hu, Nikhil Ravi, and Cynthia Rudin.
\newblock Pulse: Self-supervised photo upsampling via latent space exploration
  of generative models.
\newblock In \emph{Proceedings of the IEEE/CVF Conference on Computer Vision
  and Pattern Recognition (CVPR)}, June 2020.

\bibitem[Mescheder et~al.(2018)Mescheder, Geiger, and
  Nowozin]{MeschederICML2018}
Lars Mescheder, Andreas Geiger, and Sebastian Nowozin.
\newblock Which training methods for gans do actually converge?
\newblock In \emph{International Conference on Machine learning (ICML)}, 2018.

\bibitem[Metz et~al.(2016)Metz, Poole, Pfau, and
  Sohl-Dickstein]{metz2016unrolled}
Luke Metz, Ben Poole, David Pfau, and Jascha Sohl-Dickstein.
\newblock Unrolled generative adversarial networks.
\newblock \emph{arXiv preprint arXiv:1611.02163}, 2016.

\bibitem[Miyato et~al.(2018)Miyato, Kataoka, Koyama, and
  Yoshida]{miyato2018spectral}
Takeru Miyato, Toshiki Kataoka, Masanori Koyama, and Yuichi Yoshida.
\newblock Spectral normalization for generative adversarial networks.
\newblock \emph{arXiv preprint arXiv:1802.05957}, 2018.

\bibitem[Mordido et~al.(2020)Mordido, Yang, and
  Meinel]{mordido2020microbatchgan}
Gon{\c{c}}alo Mordido, Haojin Yang, and Christoph Meinel.
\newblock microbatchgan: Stimulating diversity with multi-adversarial
  discrimination.
\newblock \emph{arXiv preprint arXiv:2001.03376}, 2020.

\bibitem[Nguyen et~al.(2017)Nguyen, Le, Vu, and Phung]{nguyen2017dual}
Tu~Nguyen, Trung Le, Hung Vu, and Dinh Phung.
\newblock Dual discriminator generative adversarial nets.
\newblock In \emph{Advances in Neural Information Processing Systems}, pages
  2670--2680, 2017.

\bibitem[Nie et~al.(2018)Nie, Trullo, Lian, Wang, Petitjean, Ruan, Wang, and
  Shen]{nie2018medical}
Dong Nie, Roger Trullo, Jun Lian, Li~Wang, Caroline Petitjean, Su~Ruan, Qian
  Wang, and Dinggang Shen.
\newblock Medical image synthesis with deep convolutional adversarial networks.
\newblock \emph{IEEE Transactions on Biomedical Engineering}, 65\penalty0
  (12):\penalty0 2720--2730, 2018.

\bibitem[Radford et~al.(2015)Radford, Metz, and
  Chintala]{radford2015unsupervised}
Alec Radford, Luke Metz, and Soumith Chintala.
\newblock Unsupervised representation learning with deep convolutional
  generative adversarial networks.
\newblock \emph{arXiv preprint arXiv:1511.06434}, 2015.

\bibitem[Rajasegaran et~al.(2019)Rajasegaran, Hayat, Khan, Khan, and
  Shao]{NIPS2019_9429}
Jathushan Rajasegaran, Munawar Hayat, Salman~H Khan, Fahad~Shahbaz Khan, and
  Ling Shao.
\newblock Random path selection for continual learning.
\newblock In \emph{Advances in Neural Information Processing Systems 32}, pages
  12669--12679. Curran Associates, Inc., 2019.

\bibitem[Rao et~al.(2020)Rao, Harris, Irpan, Levine, Ibarz, and
  Khansari]{rao2020rl}
Kanishka Rao, Chris Harris, Alex Irpan, Sergey Levine, Julian Ibarz, and Mohi
  Khansari.
\newblock Rl-cyclegan: Reinforcement learning aware simulation-to-real.
\newblock In \emph{Proceedings of the IEEE/CVF Conference on Computer Vision
  and Pattern Recognition}, pages 11157--11166, 2020.

\bibitem[Ratcliff(1990)]{ratcliff1990connectionist}
Roger Ratcliff.
\newblock Connectionist models of recognition memory: constraints imposed by
  learning and forgetting functions.
\newblock \emph{Psychological review}, 97\penalty0 (2):\penalty0 285, 1990.

\bibitem[Rebuffi et~al.(2017)Rebuffi, Kolesnikov, Sperl, and
  Lampert]{rebuffi2017icarl}
Sylvestre-Alvise Rebuffi, Alexander Kolesnikov, Georg Sperl, and Christoph~H
  Lampert.
\newblock icarl: Incremental classifier and representation learning.
\newblock In \emph{Proceedings of the IEEE conference on Computer Vision and
  Pattern Recognition}, pages 2001--2010, 2017.

\bibitem[Rusu et~al.(2016)Rusu, Rabinowitz, Desjardins, Soyer, Kirkpatrick,
  Kavukcuoglu, Pascanu, and Hadsell]{rusu2016progressive}
Andrei~A Rusu, Neil~C Rabinowitz, Guillaume Desjardins, Hubert Soyer, James
  Kirkpatrick, Koray Kavukcuoglu, Razvan Pascanu, and Raia Hadsell.
\newblock Progressive neural networks.
\newblock \emph{arXiv preprint arXiv:1606.04671}, 2016.

\bibitem[Salimans et~al.(2016)Salimans, Goodfellow, Zaremba, Cheung, Radford,
  and Chen]{salimans2016improved}
Tim Salimans, Ian Goodfellow, Wojciech Zaremba, Vicki Cheung, Alec Radford, and
  Xi~Chen.
\newblock Improved techniques for training gans, 2016.

\bibitem[Srivastava et~al.(2017)Srivastava, Valkov, Russell, Gutmann, and
  Sutton]{srivastava2017veegan}
Akash Srivastava, Lazar Valkov, Chris Russell, Michael~U Gutmann, and Charles
  Sutton.
\newblock Veegan: Reducing mode collapse in gans using implicit variational
  learning.
\newblock In \emph{Advances in Neural Information Processing Systems}, pages
  3308--3318, 2017.

\bibitem[Thanh-Tung and Tran(2020)]{thanhcatastrophic}
Hoang Thanh-Tung and Truyen Tran.
\newblock On catastrophic forgetting and mode collapse in gans.
\newblock \emph{arXiv preprint arXiv:1807.04015}, 2020.

\bibitem[Tolstikhin et~al.(2017)Tolstikhin, Gelly, Bousquet, Simon-Gabriel, and
  Sch{\"o}lkopf]{tolstikhin2017adagan}
Ilya~O Tolstikhin, Sylvain Gelly, Olivier Bousquet, Carl-Johann Simon-Gabriel,
  and Bernhard Sch{\"o}lkopf.
\newblock Adagan: Boosting generative models.
\newblock In \emph{Advances in neural information processing systems}, pages
  5424--5433, 2017.

\bibitem[Tran et~al.(2019)Tran, Tran, Nguyen, Yang, et~al.]{tran2019self}
Ngoc-Trung Tran, Viet-Hung Tran, Bao-Ngoc Nguyen, Linxiao Yang, et~al.
\newblock Self-supervised gan: Analysis and improvement with multi-class
  minimax game.
\newblock In \emph{Advances in Neural Information Processing Systems}, pages
  13253--13264, 2019.

\bibitem[Welinder et~al.(2010)Welinder, Branson, Mita, Wah, Schroff, Belongie,
  and Perona]{WelinderEtal2010}
P.~Welinder, S.~Branson, T.~Mita, C.~Wah, F.~Schroff, S.~Belongie, and
  P.~Perona.
\newblock {Caltech-UCSD Birds 200}.
\newblock Technical Report CNS-TR-2010-001, California Institute of Technology,
  2010.

\bibitem[Wolterink et~al.(2017)Wolterink, Leiner, Viergever, and
  I{\v{s}}gum]{wolterink2017generative}
Jelmer~M Wolterink, Tim Leiner, Max~A Viergever, and Ivana I{\v{s}}gum.
\newblock Generative adversarial networks for noise reduction in low-dose ct.
\newblock \emph{IEEE transactions on medical imaging}, 36\penalty0
  (12):\penalty0 2536--2545, 2017.

\bibitem[Yi et~al.(2019)Yi, Walia, and Babyn]{yi2019generative}
Xin Yi, Ekta Walia, and Paul Babyn.
\newblock Generative adversarial network in medical imaging: A review.
\newblock \emph{Medical image analysis}, 58:\penalty0 101552, 2019.

\bibitem[Zhang et~al.(2019)Zhang, Sax, Zamir, Guibas, and
  Malik]{sidetuning2019}
Jeffrey~O. Zhang, Alexander Sax, Amir Zamir, Leonidas~J. Guibas, and Jitendra
  Malik.
\newblock Side-tuning: Network adaptation via additive side networks.
\newblock 2019.

\bibitem[Zhang et~al.(2018)Zhang, Zhang, Zhang, Liu, and
  Khurshid]{zhang2018deeproad}
Mengshi Zhang, Yuqun Zhang, Lingming Zhang, Cong Liu, and Sarfraz Khurshid.
\newblock Deeproad: Gan-based metamorphic testing and input validation
  framework for autonomous driving systems.
\newblock In \emph{2018 33rd IEEE/ACM International Conference on Automated
  Software Engineering (ASE)}, pages 132--142. IEEE, 2018.

\end{thebibliography}

\end{document}